\documentclass[conference]{IEEEtran}
\IEEEoverridecommandlockouts
\usepackage{cite}
\usepackage{amsmath,amssymb,amsfonts}
\usepackage{algorithmic}
\usepackage{graphicx}
\usepackage{textcomp}
\usepackage{xcolor}
\usepackage{mathtools}
\usepackage{bbm}

\def\BibTeX{{\rm B\kern-.05em{\sc i\kern-.025em b}\kern-.08em
    T\kern-.1667em\lower.7ex\hbox{E}\kern-.125emX}}
    
\newcommand\independent{\protect\mathpalette{\protect\independenT}{\perp}}
\def\independenT#1#2{\mathrel{\rlap{$#1#2$}\mkern2mu{#1#2}}}

\newcommand*\diff{\mathop{}\!\mathrm{d}}

\begin{document}

\title{Temporal Registration in Application to \\ In-utero MRI Time Series
\thanks{Ruizhi~Liao (email: ruizhi@mit.edu), Elfar~Adalsteinsson, and Polina~Golland are with Massachusetts Institute of Technology, Cambridge, MA, USA.}
\thanks{Esra~A.~Turk and P.~Ellen~Grant are with Boston Children's Hospital, Harvard Medical School, Boston, MA, USA.}
\thanks{Miaomiao~Zhang is with Washington University in St. Louis, MO, USA.}
\thanks{Jie~Luo is with Shanghai Jiao Tong University, Shanghai, China.}
\thanks{This work has been submitted to the IEEE-Transactions on Medical Imaging for possible publication. Copyright may be transferred without notice, after which this version may no longer be accessible.}
\thanks{This work was supported in part by NIH NIBIB NAC P41EB015902, NIH NICHD U01HD087211, NIH NIBIB R01EB017337, Shanghai Sailing Program 18YF1410900, Wistron Corporation, and Merrill Lynch Fellowship.}
}

\author{\IEEEauthorblockN{Ruizhi~Liao, Esra~A.~Turk, Miaomiao~Zhang, Jie~Luo\\Elfar~Adalsteinsson, P.~Ellen~Grant, Polina~Golland}}

\maketitle

\begin{abstract}
We present a robust method to correct for motion in volumetric in-utero MRI time series. Time-course analysis for in-utero volumetric MRI time series often suffers from substantial and unpredictable fetal motion. Registration provides voxel correspondences between images and is commonly employed for motion correction. Current registration methods often fail when aligning images that are substantially different from a template (reference image). To achieve accurate and robust alignment, we make a Markov assumption on the nature of motion and take advantage of the temporal smoothness in the image data. Forward message passing in the corresponding hidden Markov model (HMM) yields an estimation algorithm that only has to account for relatively small motion between consecutive frames. We evaluate the utility of the temporal model in the context of in-utero MRI time series alignment by examining the accuracy of propagated segmentation label maps. Our results suggest that the proposed model captures accurately the temporal dynamics of transformations in in-utero MRI time series. 
\end{abstract}


\section{Introduction}
This paper demonstrates a robust temporal image registration framework. Large motion presents significant challenges for time-course analysis in dynamic image series. To correct for motion in medical images, registration that produces voxel correspondences is commonly employed. Unfortunately, registration often fails if the motion is substantial. We take advantage of the temporal smoothness of motion to achieve robust alignment of volumetric MRI time series. This work is motivated by a clinical research application in monitoring placental and fetal health, which we introduce below.

\begin{figure*}[h]
\centerline{
\hfill
\includegraphics[width=6.6 in]{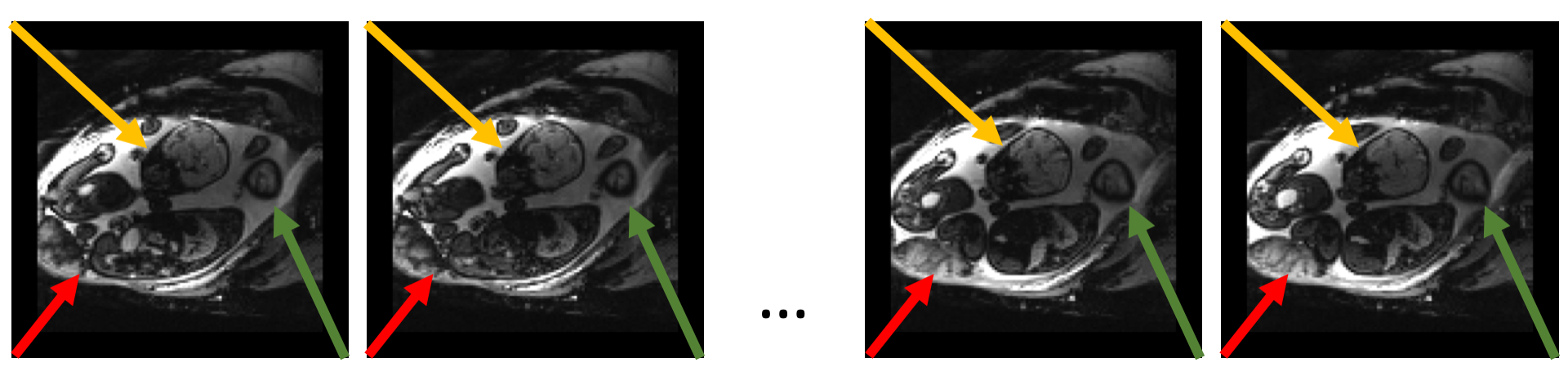}
\hfill
}
\centerline{
\hfill
\begin{minipage}[t]{1.35in}
\centering
$1$
\end{minipage} 
\hskip0.1in
\begin{minipage}[t]{1.35in}
\centering
$2$
\end{minipage} 
\hskip0.55in
\begin{minipage}[t]{1.35in}
\centering
$124$
\end{minipage} 
\hskip0.1in
\begin{minipage}[t]{1.35in}
\centering
$125$
\end{minipage} 
\hfill
}
\caption{Example twin pregnancy case from the study. The same 2D cross-section from volumetric frames~$1$, $2$, $124$, and $125$ is shown. Arrows indicate areas of substantial motion of the fetal heads (green and yellow) and highly non-rigid motion of the placenta (red). }
\label{fig:inutero_original}
\end{figure*}

\subsection{Motivating application}

In-utero blood oxygenation level dependent (BOLD) magnetic resonance imaging (MRI) is a promising imaging tool for studying functional dynamics of placenta and fetal organs~\cite{schopf2012watching, sorensen2013changes, sorensen2013changes2, jie2016ismrm}. Changes in fetal and placental oxygenation levels with maternal hyperoxygenation can be used for charactering and detecting placental dysfunction~\cite{aimot2013vivo}. Temporal MRI data suffers from serious motion artifacts due to maternal respiration, unpredictable fetal movement, and signal non-uniformities~\cite{studholme2011mapping}, as illustrated in Fig.~\ref{fig:inutero_original}. Different types of motion are presented in these MRI time series. The fetal brains move as rigid objects with a wide range of motion; the placenta moves more locally and deforms non-rigidly. In our study, the image series include about 500 volumes~(frames) that are analyzed to study fetal and placental physiology. 

\subsection{Related Work}

Temporal registration has been previously investigated in time series of cardiac images~\cite{sundar2009biomechanically, park1996deformable, chandrashekara2003construction, metz2011nonrigid} and lung images~\cite{metz2011nonrigid, mcclelland2013respiratory, rietzel2006deformable, reinhardt2008registration}. Both cardiac motion and breathing patterns are somewhat regular and smooth across time, enabling biomechanical modeling~\cite{sundar2009biomechanically, mcclelland2013respiratory, park1996deformable}. Electrocardiography~(ECG) gating or respiratory gating can be incorporated in the cardiac or lung image motion correction, by either enforcing the transformation models to be periodic or penalizing non-periodic transformations in the cost function. These approaches cannot be generalized to images that contain complex non-periodic motion patterns.

Existing registration techniques for dynamic medical imaging data can be categorized into two distinct groups. The first type of methods performs registration based on the differences between each pair of consecutive frames. Each pair is aligned, and the estimated transformations are concatenated to estimate alignment of all images in the series~\cite{reinhardt2008registration, boldea20084d}. In application to long image series like our data, this approach leads to substantial accumulated errors after several concatenation steps. These methods only take the moving image and the reference image into account in each pairwise registration step, and ignore essential temporal information from other images in the series. The second type applies groupwise registration to simultaneously align each individual image to a group template -- estimated or selected from one of the input images -- aiming to yield acceptable registration results for all images in the series~\cite{balci2007non, bhatia2004consistent, miller2000learning, zollei2005efficient, rietzel2006deformable, marsland2008minimum, metz2011nonrigid, durrleman2013toward, singh2015hierarchical}. Taking this approach in our application, registration fails for a great number of images that are substantially different from the template, necessitating outlier detection and rejection~\cite{turk2017spatiotemporal}. Alternatively, a global objective function based on pairwise differences between each individual image in the series and an implicitly defined template frame can incorporate the temporal information~\cite{metz2011nonrigid}. This leads to a global optimization process, i.e., the algorithm performs pairwise registration of consecutive frames iteratively until the entire series comes into alignment. This type of global optimization is extremely challenging in large image sets like our time series.

The problem of temporal alignment has also been investigated in longitudinal image analysis that aims to characterize temporal variations~\cite{davis2010population, khan2008representation, durrleman2013toward, fishbaugh2011estimation, reuter2011avoiding}. The goal is to examine the behavior of a biological system driven by development, aging, or disease processes~\cite{fishbaugh2011estimation}. The statistical analysis is performed via temporal shape regression on a Riemannian manifold~\cite{davis2010population, khan2008representation, durrleman2013toward, fishbaugh2011estimation} to construct a trajectory that characterizes the shape variation across time. Longitudinal data is often sparsely distributed in time, and temporal smoothness constraint is therefore imposed. Typically, the challenges in longitudinal analysis involve growth or degeneration, which requires modeling tools that capture subtle deformation and are distinct from those in our problem of MRI time series alignment in the presence of substantial motion. 

Unlike other applications that can require subjects to remain stationary, fetal and placental imaging is an application with irregular motion that cannot be controlled at the time of acquisition.

\begin{figure*}[h]
\centerline{
\hfill
\includegraphics[width=5in]{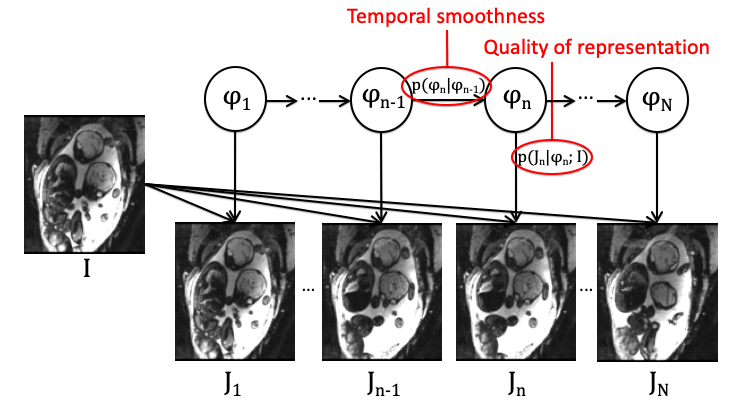}
\hfill
}
\caption{A graphical model of the temporal HMM. The likelihood term $p(J_n|\varphi_n; I)$ gives rise to an image matching term in registration, and the transition probability distribution~$p({\varphi_n}|\varphi_{n-1})$ encodes the temporal and spatial smoothness of the transformation.}
\vskip-0.1in
\label{fig:graphical_model}
\end{figure*}

\subsection{Proposed Temporal Registration Method}

To overcome the difficulty of correcting large and irregular motion in a long MRI time series, we assume a Markov structure that induces temporal smoothness and offers an efficient way to estimate the transformations by serially solving pairwise registration problems. We derive filtered estimates of the transformations from the hidden Markov model that represents the MRI series. The resulting sequential procedure estimates the transformations of the template to each frame in the series. We provide a flexible framework for temporal dense alignment of image time series.

Unlike the methods that perform pairwise registration of consecutive frames and concatenate the estimated transformations, we use temporal smoothness explicitly at every estimation step. The sequential estimation framework yields an efficient algorithm that makes alignment of long MRI series feasible. Our approach is related to filtering methods in respiratory motion modeling~\cite{mcclelland2013respiratory}. We do not model the motion explicitly but rather capture it through transformation of organs of interest to template to adapt to complex motion patterns.

We demonstrate and evaluate the proposed method in the context of the in-utero BOLD MRI study by providing registration-based tracking for organs of interest. We employ the rigid motion model for fetal brains and a non-rigid B-Spline motion model for placentae. The experimental results offer significant improvements over pairwise registration in alignment of fetal brains that tend to move substantially, and robust improvements in alignment of placentae.

This paper presents an extended version of the preliminary algorithm presented at the International Conference on Medical Image Computing and Computer-Assisted Intervention (MICCAI)~\cite{liao2016temporal}. In this paper, we derive in detail the instantiation of a hidden Markov model in the context of MRI time series. We model the motion as a latent state in the hidden Markov model, derive a temporal registration algorithm using filtered estimates, and discuss the challenges of developing a smoothing algorithm. Here we employ rigid and B-Spline transformation models on the regions of fetal brains and placentae respectively, instead of one single transformation model for the entire volumetric image used in~\cite{liao2016temporal}, to adapt to the different motion patterns in the two regions. Finally, extensive experiments investigate the sensitivity of the resulting registration on the model parameters. 

This paper is organized as follows. In the next section, we present the generative model for temporal motion and observed images using hidden Markov model, derive the resulting algorithm for filtered estimates, and provide implementation details. In Section~\ref{section:results}, we introduce the in-utero BOLD MRI data and report the experimental results. In Section~\ref{section:discussion}, we discuss future directions, methodological challenges of developing a backward message passing algorithm for temporal registration, and potential extension for other transformation models. Conclusions follow in Section~\ref{section:conclusions}. 

\section{Methods}
\label{section:methods}

We employ hidden Markov model (HMM) to represent the temporal changes in the MRI time series. One image in the MRI time series is chosen as a template, whose transformations represent the rest of the images in the series. We treat the transformation as a hidden state, and each image as an observation that depends on the template and the corresponding hidden state. In the following subsections, we present inference in HMMs using filtered estimates~\cite{baum1966statistical, bishop2006pattern} in the context of temporal registration, and describe the implementation details.

\subsection{HMM and Filtered Estimates}

The temporal dynamics of latent~(hidden) states in a HMM exhibit the Markov property, i.e., the current state depends on the history only through the previous state. Thus, inference based on the entire series of past observations is equivalent to integrating the information from the previous state and from the corresponding observation, leading to an efficient sequential estimation algorithm.

In our application, one image is chosen as the template~$I$ to be treated as a global parameter shared by all observed images. We assume that the template~$I$ deforms at each time point, and the transformation~$\varphi_n$ of the template is the latent state at time~$n\in\{1, ..., N\}$, where $N$ is the total number of images in the series. The observed image~$J_n$ at time~$n$ is generated by applying the anatomical transformation~$\varphi_n$ to the template~$I$ independently of all other time points. Fig.~\ref{fig:graphical_model} illustrates this model.

Our aim is to estimate the latent state variable~$\varphi_n$ from the observations~$J_{1:{n}}$, where we use $J_{k:m}$ and $\varphi_{k:m}$ to denote sub-series~$\{J_k, J_{k+1}, ..., J_m\}$ of the MRI time series~$\{J_1, J_2, ..., J_N\}$ and the associated transformations~$\{\varphi_k, \varphi_{k+1}, ..., \varphi_m\}$. Formally, the estimation of the latent variable~$\varphi_n$ is reduced to maximization of its posterior distribution~$p(\varphi_n|{J_{1:{n}}}; I)$ parameterized by the template~$I$. This posterior distribution, also known as the filtered estimate of the state, can be efficiently computed using forward message passing~\cite{bishop2006pattern, baum1966statistical}, also known as sequential estimation. The posterior distribution~$p(\varphi_n|{J_{1:{n}}}; I)$ is determined by integrating the previous posterior distribution~$p(\varphi_{n-1}|{J_{1:{n-1}}}; I)$ with the temporal dynamics~$p({\varphi_n}|\varphi_{n-1})$ and the likelihood~$p(J_n|\varphi_{n}; I)$ of the current observation:
\begin{align}
& p({\varphi_n}|J_{1:n}; I) \nonumber \\
& \propto  p({\varphi_n}, J_{1:{n-1}}, J_n; I) \label{forwardMessagePassing:1} \\
& \propto  p(J_n|\varphi_{n}; I) \, p({\varphi_n}|J_{1:{n-1}}; I) \label{forwardMessagePassing:2}\\
& = p(J_n|\varphi_{n}; I) \!\! \int\limits_{{{\varphi}_{{n-1}}}} \!\!\! p({\varphi_n}|\varphi_{n-1}) \, p({\varphi_{n-1}}|J_{1:{n-1}}; I)\diff{\varphi_{n-1}}, \label{forwardMessagePassing:3}
\end{align}
where Eq.~(\ref{forwardMessagePassing:2}) follows from the Markov property:
\begin{equation}
J_n{\independent}J_{1:n-1} | \varphi_n{.} \nonumber
\end{equation}
Applied recursively, the forward message passing produces the posterior distribution $p(\varphi_n|J_{1:n}; I)$ for each time point $n$ in the number of steps that is linear with $n$. Similarly, backward message passing enables inference of the posterior distribution~$p(\varphi_n|J_{1:N}; I)$, which is based on all data, often referred to as smoothing. In this paper, we incorporate temporal smoothness in temporal registration by filtering. We discuss the challenges of developing a smoothing algorithm for temporal registration in Section~\ref{section:conclusions}. 

\subsection{Estimating Temporal Transformations in HMM}

Since it is intractable to integrate over all possible transformation fields, we employ a commonly used approach of approximating the posterior distribution~$p(\varphi_n|J_{1:n}; I)$ as a point estimate that maximizes the probability distribution. In particular, if $\varphi_{n-1}^{*}$ is the best transformation estimated by the algorithm for time point $n-1$, the message from the state~$n-1$ to state~$n$ is the optimal transformation $\varphi_{n-1}^{*}$:
\begin{align}
& p({\varphi_n}|J_{1:n}; I)  \nonumber \\ 
& \propto  p(J_n|\varphi_{n}; I)
\!\! \int\limits_{{{\varphi}_{{n-1}}}} \!\!\! p({\varphi_n}|\varphi_{n-1}) \, p({\varphi_{n-1}}|J_{1:{n-1}}; I)\diff{\varphi_{n-1}} \nonumber \\
& \approx p(J_n|\varphi_{n}; I) \!\!\int\limits_{\varphi_{n-1}}  \!\!\! p({\varphi_n}|\varphi_{n-1}) \, \mathbbm{1}\{\varphi_{n-1}=\varphi_{n-1}^{*}\}\diff{\varphi_{n-1}} \nonumber \\
& = p(J_n|\varphi_{n}; I)\ p({\varphi_{n}}|\varphi_{n-1}^{*}), \label{messagePassing}
\end{align}
yielding the (recursive) estimate of the hidden state at time point $n$:
\begin{align}
\varphi_{n}^{*} & = \arg\max_{\varphi_{n}}\ p({\varphi_n}|J_{1:n}; I) \nonumber \\
& \approx \arg\max_{\varphi_{n}}\ p(J_{n}|\varphi_{n}; I)
\, p({\varphi_{n}}|\varphi_{n-1}^{*}) \label{max:def}.
\end{align}
The estimate~$\varphi_{n}^{*}$ is then used to compute the optimal transformation~$\varphi_{n+1}^{*}$ at time~$n+1$, and so on until we reach the end of the time series. 

\begin{figure*}[h]
\centerline{
\hfill
\includegraphics[width=1.7 in]{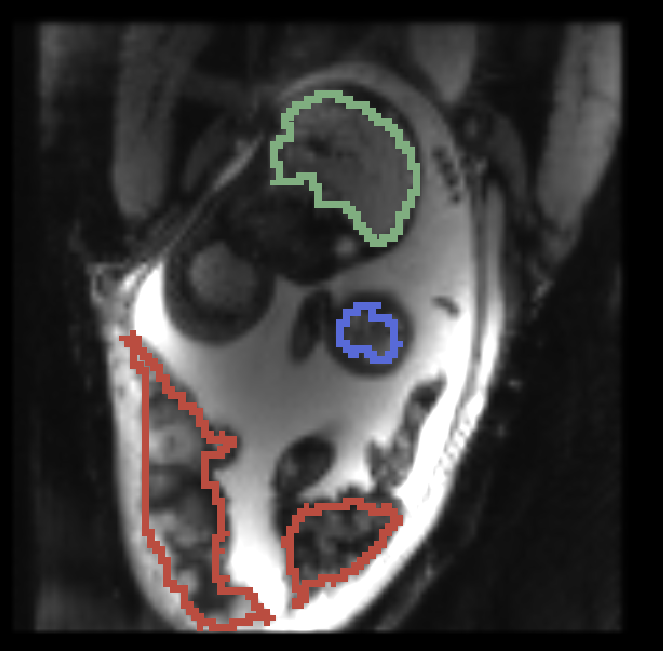}
\hskip0.02in
\includegraphics[width=1.7 in]{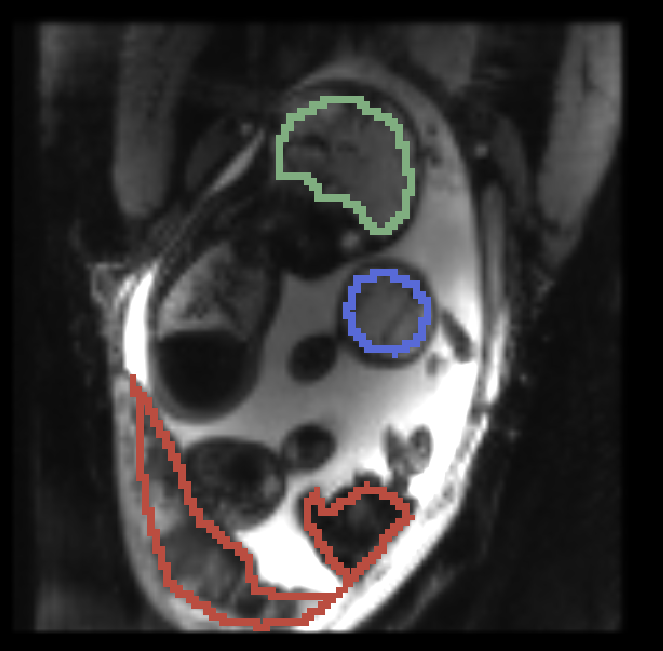}
\hskip0.02in
\includegraphics[width=1.7 in]{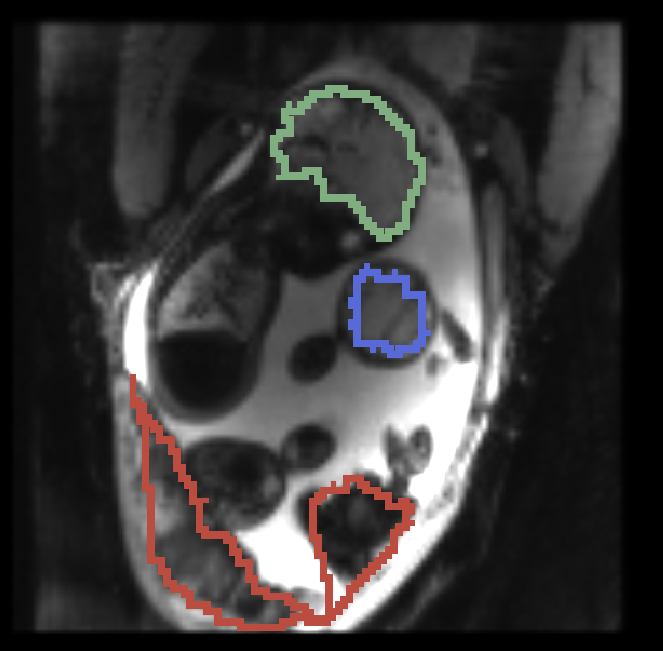}
\hfill
}
\centerline{
\hfill
\begin{minipage}[t]{1.7in}
\centering
$I=J_{250}$, \\ manual segmentation on $J_{250}$
\end{minipage} 
\hskip0.05in
\begin{minipage}[t]{1.7in}
\centering
$J_1$, \\ manual segmentation on $J_1$
\end{minipage} 
\hskip0.05in
\begin{minipage}[t]{1.75in}
\centering
$J_1$, \\ propagated segmentation
\end{minipage} 
\hfill
}
\vskip0.1in
\centerline{
\hfill
\includegraphics[width=1.7 in]{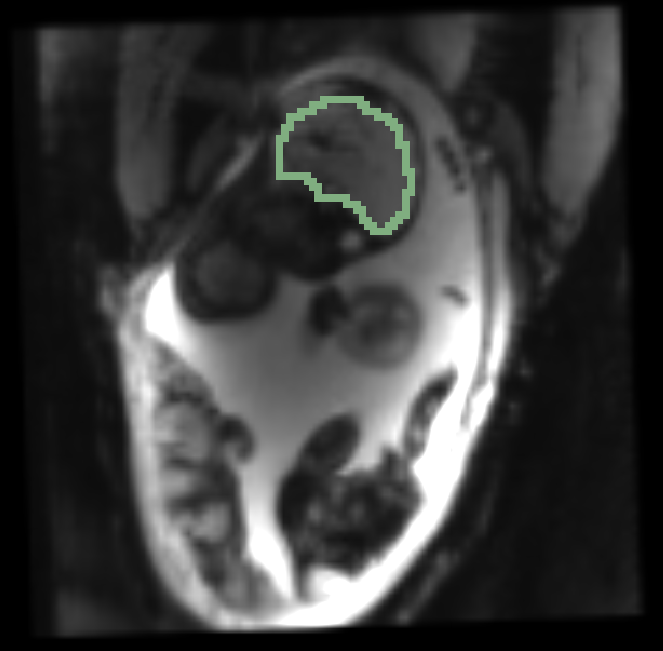}
\hskip0.02in
\includegraphics[width=1.7 in]{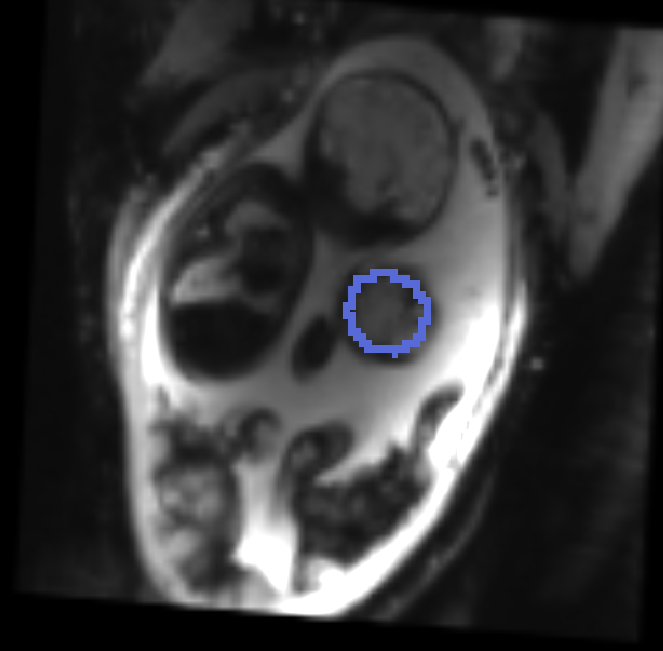}
\hskip0.02in
\includegraphics[width=1.7 in]{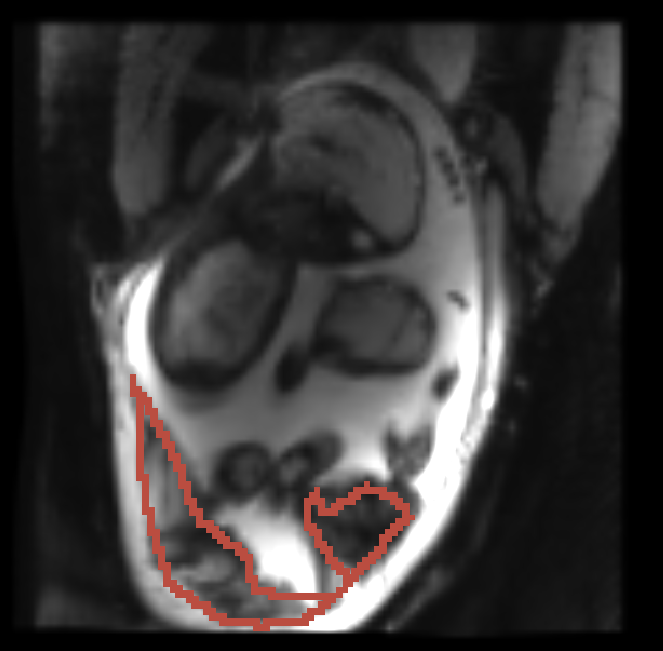}
\hfill
}
\centerline{
\hfill
\begin{minipage}[t]{1.7in}
\centering
$I(\varphi^{-1}_{250})$, \\ manual segmentation on $J_1$
\end{minipage} 
\hskip0.05in
\begin{minipage}[t]{1.7in}
\centering
$I(\varphi^{-1}_{250})$, \\ manual segmentation on $J_1$
\end{minipage} 
\hskip0.05in
\begin{minipage}[t]{1.7in}
\centering
$I(\varphi^{-1}_{250})$, \\ manual segmentation on $J_1$
\end{minipage} 
\hfill
}
\caption{Example twin pregnancy case from the study. Top row:
  Reference frame~$I=J_{250}$ with manual segmentations, frame~$J_{1}$ with manual segmentations, 
  and frame~$J_{1}$ with segmentations propagated from the manual segmentations on $I$ using the transformation estimates when regions of fetal brains and placenta are respectively counted for. Segmentations of fetal brains (green and blue) and placenta (red) are shown. 
  Botton row:
  the reference frame~$I(\varphi^{-1}_{250})$ transformed into the coordinate
  system of frame $J_{1}$ when regions of fetal brains and placenta are respectively counted for. 
  Two-dimensional cross-sections are used for visualization purposes only; all computations are performed
  in 3D.  }
\vskip-0.1in
\label{fig:labels_transferred}
\end{figure*}

\subsection{Model Setup}
\label{subsection_model_setup}

We model the likelihood term $p(J_n|\varphi_n; I)$ in Eq.~(\ref{max:def}), also known as emission probability, as the exponential likelihood on image similarity: 
\begin{equation}
\label{imageMatching} 
p(J_n|\varphi_n; I) \propto \exp\{-\text{Dist}\left(J_n , 
I \left(\varphi^{-1}_n\right)\right)\},
\end{equation}
where $\text{Dist}(\cdot, \cdot)$ is a measure of dissimilarity~(distance) between images. This probability measures how different the transformed template and the currently observed image are given the template~$I$ and the transformation~$\varphi_n$. 

The state transition probability $p({\varphi_n}|\varphi_{n-1})$ in~Eq.~(\ref{max:def}) encourages temporal smoothness of transformations and spatial smoothness of $\varphi_n$:
\begin{equation}
  p({\varphi_n}|\varphi_{n-1}) \propto \exp\{ -\lambda_1 {\left\lVert{\varphi_n\circ\varphi_{n-1}^{-1}}\right\rVert}^2
  -\lambda_2 \text{Reg}{(\varphi_n)}\},
\end{equation}
where $\left\lVert{\cdot}\right\rVert$ is an appropriate norm that encourages transformation $\varphi_n$ to be similar to~$\varphi_{n-1}$, and $\text{Reg}{(\cdot)}$ is the regularization term that encourages spatial smoothness and other desired properties. The regularization term~$\text{Reg}{(\varphi_n)}$ is often used in image registration to restrict the estimated motion to a specific transformation group.  $\lambda_1$ and $\lambda_2$ are regularization parameters that control the amount of temporal and spatial smoothness. 

We manipulate Eq.~(\ref{max:def}) to obtain
\begin{align}
\varphi_{n}^{*} &\approx \arg\max_{\varphi_{n}}\ p(J_n|\varphi_n; I) \, 
p(\varphi_n|\varphi_{n-1}^{*})  \nonumber \\ 
\begin{split}
& = \arg\min_{\varphi_{n}}\ \text{Dist}\left(J_n ,
  I\left(\varphi^{-1}_n\right)\right) \\
  & + \lambda_1 {\left\lVert{\varphi_n\circ{(\varphi_{n-1}^{*})}^{-1}}\right\rVert}^2
+ \lambda_2 \text{Reg}{(\varphi_n)} \label{messagePassing2}
\end{split}
\end{align}
and observe that this optimization problem is reduced to serial pairwise image registration of the template $I$ and the observed image $J_n$. 


The algorithm proceeds as follows. Given the transformation estimate $\varphi_{n-1}^{*}$ of the template~$I$ to the image~$J_{n-1}$, we apply the registration algorithm to $I$ and $J_n$ by minimizing the cost function in Eq.~(\ref{messagePassing2}), while also using $\varphi_{n-1}^{*}$ as an initialization, resulting in the estimate~$\varphi_{n}^{*}$.


This HMM-based temporal transformation model can be easily generalized to different types of similarity metrics between moving images and fixed images, temporal regularization, and transformation models. The model can be readily augmented to account for signal non-uniformities and a latent template that is estimated jointly with the transformations, similar to prior work in groupwise registration~\cite{metz2011nonrigid, durrleman2013toward, singh2015hierarchical,rietzel2006deformable}. In the following subsection, we describe our implementation for aligning in-utero BOLD MRI time series.

\begin{figure}[h!]
\centerline{
\hfill
\includegraphics[width=3.65 in, height=2.45in]{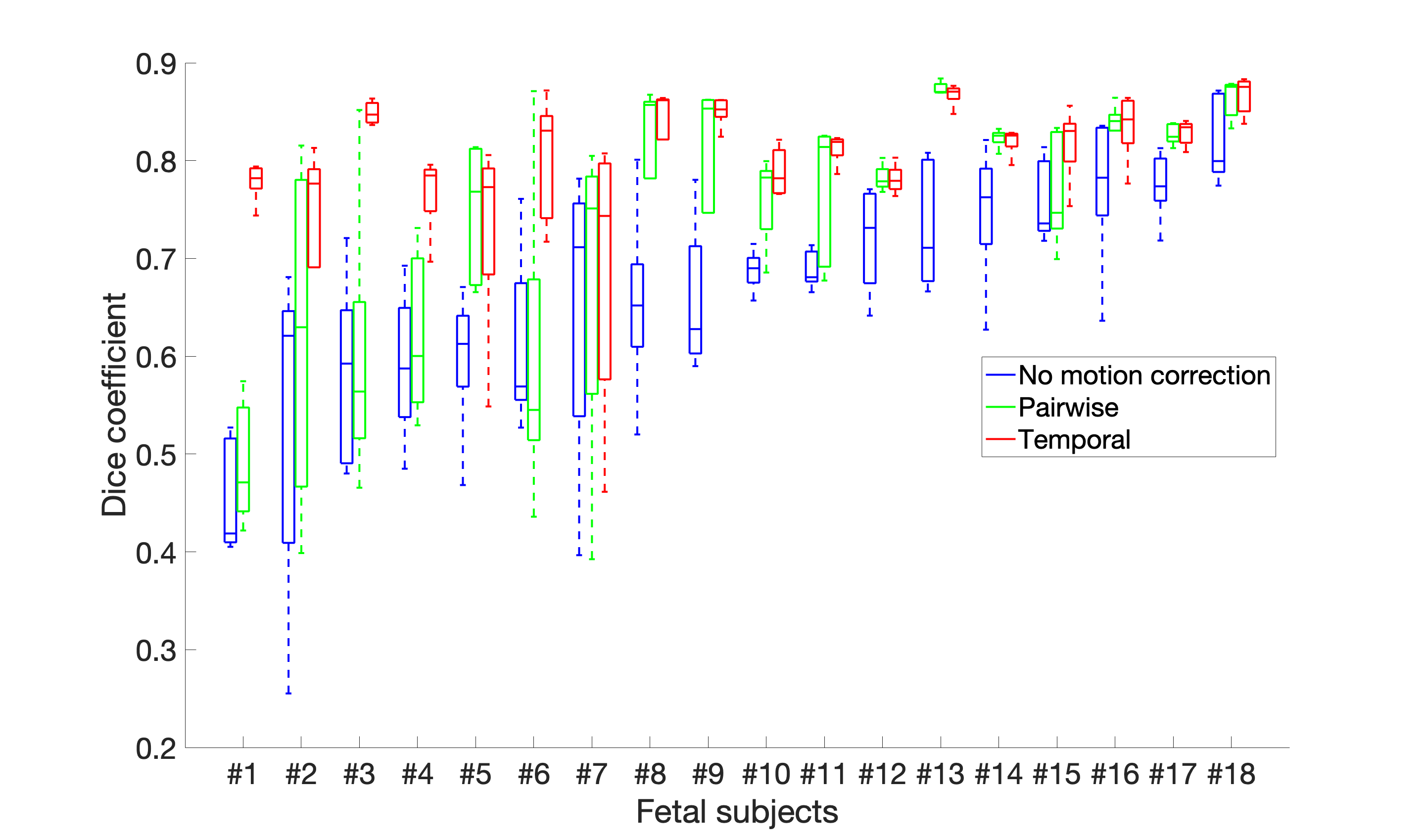}
\hfill
}
\vskip0.05in
\centerline{
\hfill
\includegraphics[width=1.75in, height=1.6in]{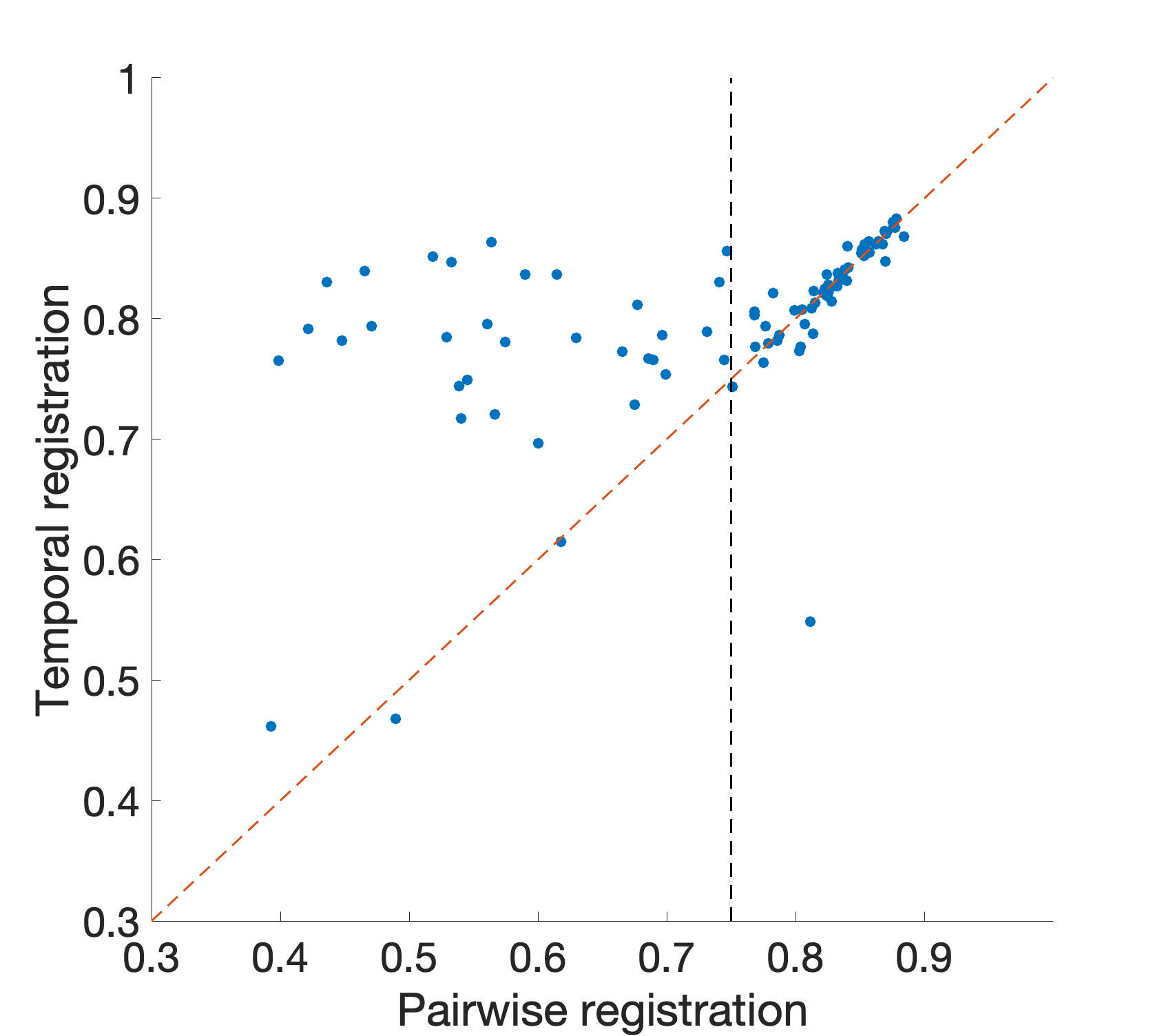}
\includegraphics[width=1.75in, height=1.6in]{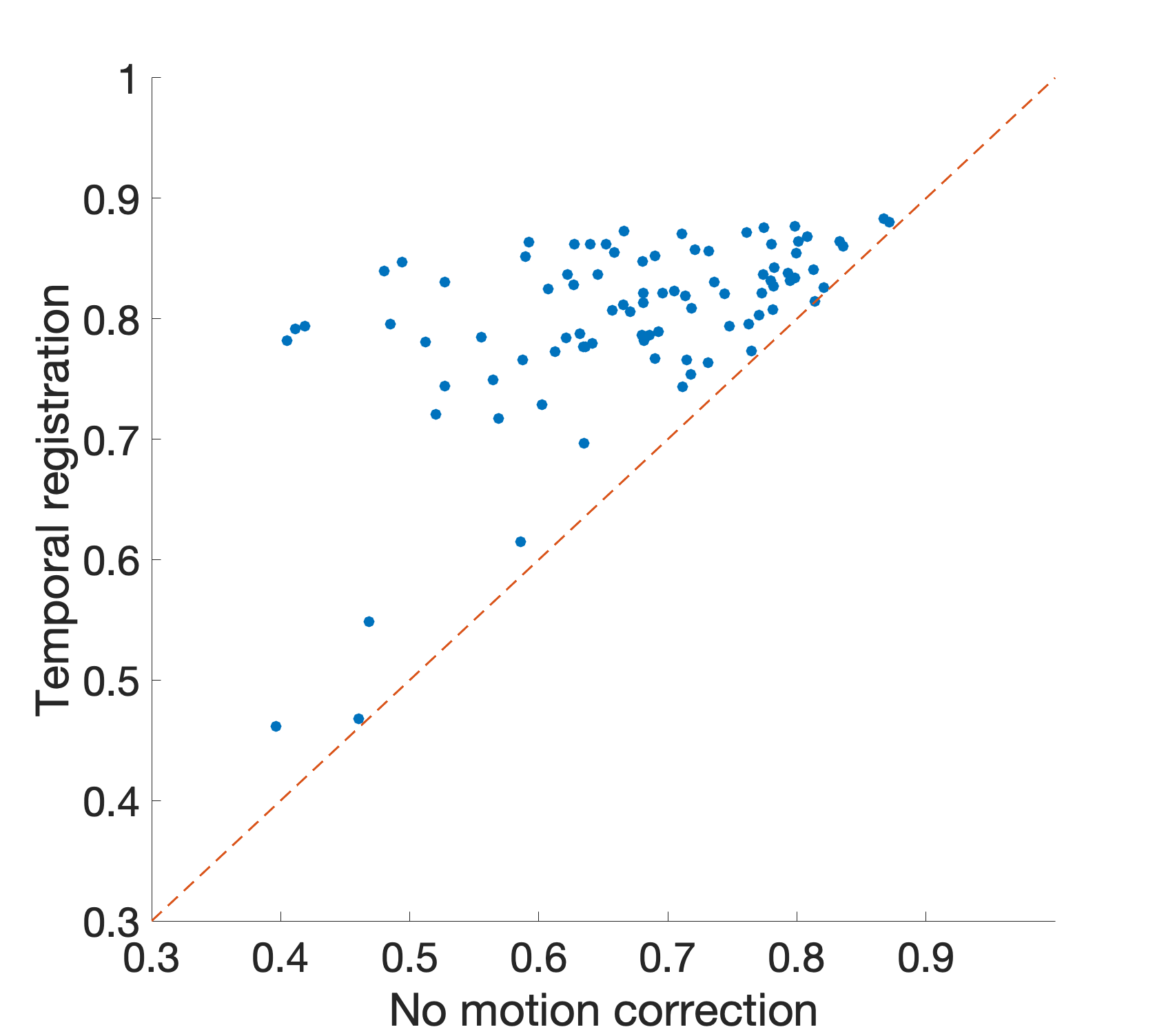}
\hfill
}
\vskip0.2in
\centerline{
\hfill
(a) Fetal brains (18 fetuses): volume overlap
\hfill
}
\vskip0.1in
\centerline{
\hfill
\includegraphics[width=3.6in, height=2.1in]{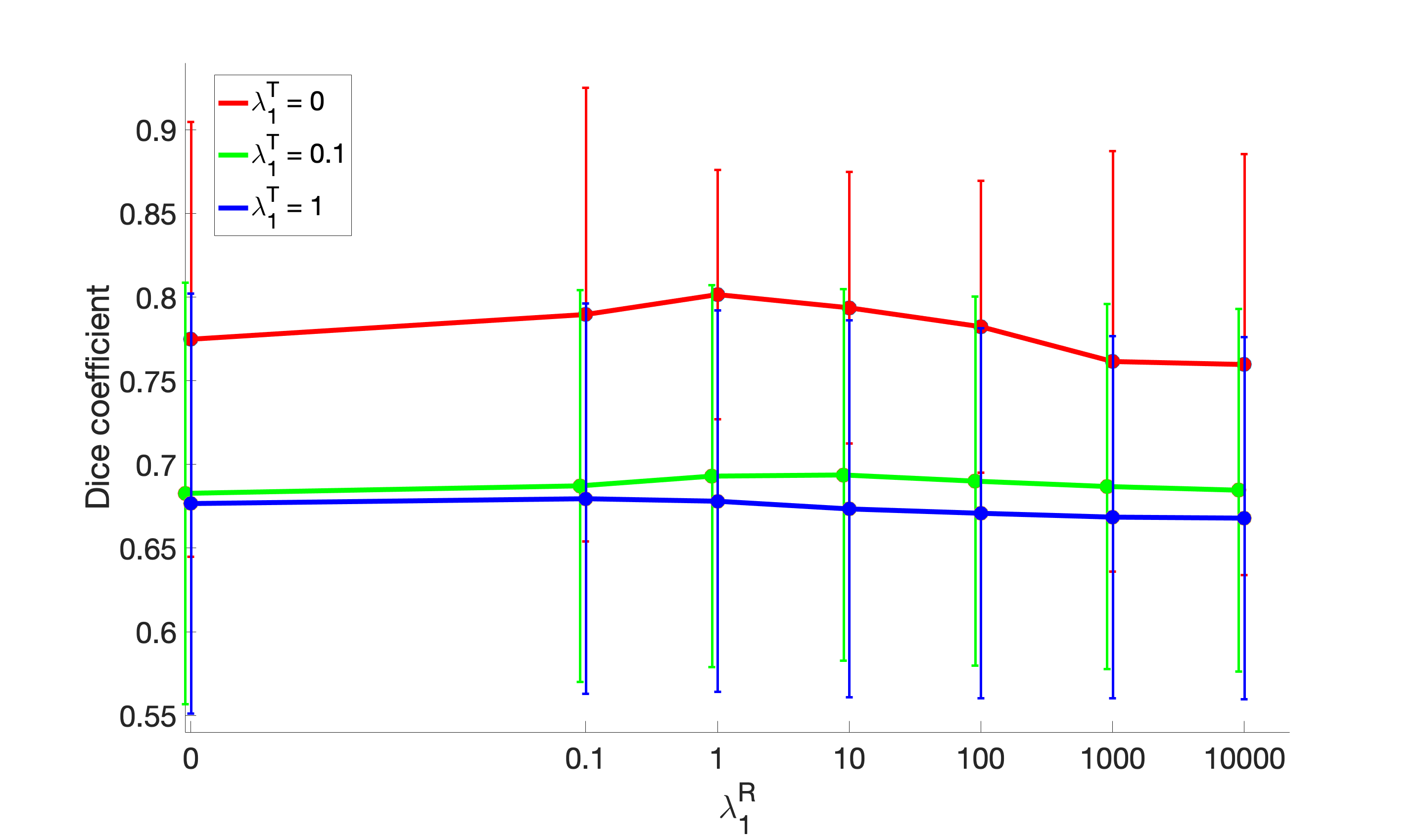}
\hfill
}
\vskip0.05in
\centerline{
\hfill
(b) Fetal brains: parameter sensitivity 
\hfill
}
\vskip0.05in
\caption{Volume overlap (Dice coefficient) between propagated and manual fetal brain segmentations. Subfigure (a): The top row shows statistics of 18 fetuses for our method (red), pairwise registration to the template frame (green), and no motion correction (blue). The 18 fetuses in the study are reported in the increasing order of volume overlap of original segmentation labels (no motion correction). The two plots in the bottom row show the Dice coefficients of the 90 images from temporal registration and pairwise registration and Dice coefficients from temporal registration and no motion correction. The volume overlap for temporal registration is $9.0\% ({\pm}16.4\%)$ on average higher than that from pairwise registration. In subjects with the pairwise volume overlap below 0.75 (34 out of 90), the average volume overlap improvement is $31.2\% ({\pm}20.9\%)$. Subfigure (b): Volume overlap as a function of the weights of temporal regularization. The plot shows the volume overlap in the region of the fetal brain with rigid transformation model as a function of $\lambda^{R}_1$, when $\lambda^{T}_1=0$ (red), $\lambda^{T}_1=0.1$ (green), and $\lambda^{T}_1=1$ (blue).
}
\label{fig:dice_brains}
\end{figure}

\subsection{Implementation Details}
\label{subsection_implementation}

We implemented our method using the Insight Segmentation and Registration Toolkit (ITK)~\cite{yoo2002engineering}. Negative cross-correlation~\cite{avants2008symmetric} is employed as the metric of image dissimilarity $\text{Dist}(\cdot, \cdot)$ as cross-correlation can handle images with signal non-uniformities that are present in the in-utero BOLD MRI data. Cross-correlation is computed over a 5$\times$5$\times$5~voxel volumetric window. We implemented the rigid transformation model to account for the fetal brain motion and the B-Spline transformation model to characterize the placental transformation. We apply rigid transformation model for fetal brain alignment by defining a region-of-interest~(ROI) mask in the template that is used to compute the registration cost function. For placentae, the B-Spline transformation model corrects the non-rigid deformation in addition to motion. For the B-Spline model, the grid spacing was set to be 10$\times$10$\times$10 voxels, in the volume of 150$\times$150$\times$120 voxels of our MRI data.

Our choice of the transformation model offers a natural implementation of spatial smoothness. For the rigid model, the third term in Eq.~(\ref{messagePassing2}) disappears, and for B-Spline model, it becomes a sum-of-squares of the B-Spline coefficients. The temporal regularization is implemented as the $L_2$ norm of the difference between the current transformation and the previously estimated transformation parameters. Specifically, the six parameters in 3-D rigid transformation model are split into two groups: three parameters representing rotation angles and three parameters representing translation. The weight of the second term in Eq.~(\ref{messagePassing2}) is then represented by $\lambda^{R}_1$ and $\lambda^{T}_1$ respectively for the two groups of parameters. $\lambda^{R}_1$ and $\lambda^{T}_1$ are set to $1$ and $0$ in the rigid registration. $\lambda_{1}$ is set to $0.005$ in the B-Spline registration.

Note that the template can be any image in the series, and the algorithm can proceed in either ascending or descending order of the series. In our experiments, we chose the image in the middle of each series as the template $I=J_{250}$. In each series, we register images $J_{1:249}$ to the template $J_{250}$ in a descending order, and register images $J_{251:N}$ to the template $J_{250}$ in an ascending order.

\section{Experimental Results}
\label{section:results}

We evaluated our method on in-utero BOLD MRI time series acquired as part of a placental function study~\cite{luo2017vivo}.

\subsection{Data}

Ten pregnant women were consented and scanned on a 3T Skyra Siemens scanner (single-shot GRE-EPI, $3 \times 3 {\text{mm}}^2$ in-plane resolution, $3 \text{mm}$ slice thickness, interleaved slice acquisition, TR$=5.8-8 \text{s}$, TE$=32-36 \text{ms}$, FA$=90^o$) using 18-channel body and 12-channel spine receive arrays. Each series contains around 300 volumetric images. This study included three singleton pregnancies, six twin pregnancies, and one triplet pregnancy, between 28 and 37 weeks of gestational age.

To eliminate the effects of slice interleaving, we resampled odd and even slices of each volumetric image onto a common isotropic~$3$mm$^3$ image grid. We split each interleaved volumetric image into even and odd slices, and linearly interpolated the acquired slices to complete the volume. The number of images in each MRI series was thus doubled to around 600. To enable quantitative evaluation, we manually delineated the fetal brains (total of 18) and placentae (total of 10) in six frames ($1$, $50$, $100$, $150$, $200$, $250$) in each series.

\subsection{Evaluation}

To evaluate the temporal model, we compare it to a variant of our method that does not assume the temporal structure and instead aligns each image in the series to the reference frame using the same registration algorithm used by our method. This approach is commonly used for fetal MRI series alignment~\cite{turk2017spatiotemporal, you2016robust}. Algorithmically, this corresponds to setting ${\lambda_1}=0$ in Eq.~(\ref{messagePassing2}) and initializing the registration step with an identity transformation instead of the previously estimated transformation~$\varphi_{n-1}^*$. The other parameter settings in the temporal model are the same with those in the pairwise model.

To quantify the accuracy of the alignment, we transform the manual segmentation labels (fetal brains and placentae) in the template~$I=J_{250}$ to the five segmented frames ($J_{1}, J_{50}, J_{100}, J_{150}, J_{200}$) in each series using the estimated transformations. We employ Dice coefficient~\cite{dice1945measures} to quantify volume overlap between the transferred and the manual segmentations. In our application, the goal is to study the average temporal signal for each ROI, and therefore delineation of an ROI provides an appropriate evaluation target. In addition, we also investigate how the alignment accuracy changes with the weight $\lambda_1$ of the temporal smoothness term in Eq.~(\ref{messagePassing2}). For B-Spline model, we additionally evaluate the log determinants of Jacobian of the estimated transformation fields.

\subsection{Results}

Fig.~\ref{fig:labels_transferred} provides example results from the study for fetal brains and placentae. We observe that the reference frame~$J_{250}$  is warped accurately by the temporal registration algorithm in the regions of fetal brains and placentae to represent the first frame~$J_{1}$ in the series that is substantially different from the template. The delineations achieved by transferring manual segmentation labels from the reference frame to the coordinate system of the selected frame are in good alignment with the manual segmentation outlines for that selected frame.

Fig.~\ref{fig:dice_brains} reports volume overlap statistics (Dice coefficient) for the fetal brains in this study. We observe that temporal alignment improves volume overlap in important ROIs over the baseline pairwise registration. We note that temporal alignment offers particularly substantial gains in cases with a lot of motion, i.e., low original volume overlap, in fetal brains. Fig.~\ref{fig:dice_brains} also shows how alignment accuracy changes with the weights of temporal smoothness regularization in Eq.~(\ref{messagePassing2}). For the rigid model, the Dice coefficients are more sensitive to $\lambda^{T}_1$, and the registration achieves the best accuracy for ${\lambda^{T}_1}=0$. Note that the temporal smoothness of translation is still somewhat maintained by initializing the current registration with the previous transformation estimation, even if ${\lambda^{T}_1}=0$. In our tests, the temporal rigid registration yields the best result when ${\lambda^{T}_1}={0}$ and ${\lambda^{R}_1}={1}$ for fetal brains.

\begin{figure}[h!]
\centerline{
\hfill
\includegraphics[width=3.7in, height=2.1in]{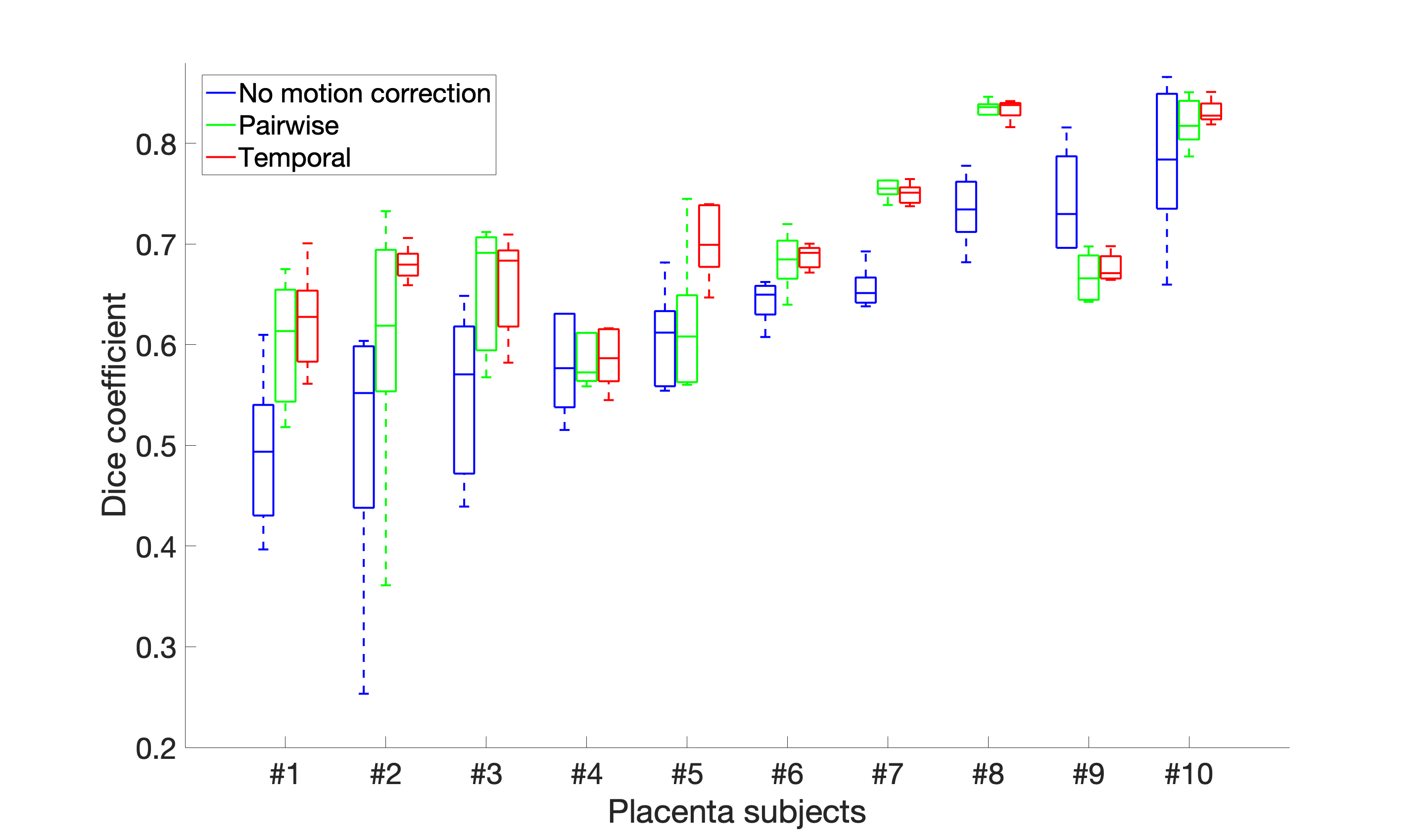}
\hfill
}
\vskip0.03in
\centerline{
\hfill
(a) Placentae (10 subjects): volume overlap
\hfill
}
\vskip0.08in
\centerline{
\hfill
\includegraphics[width=3.7in, height=2.1in]{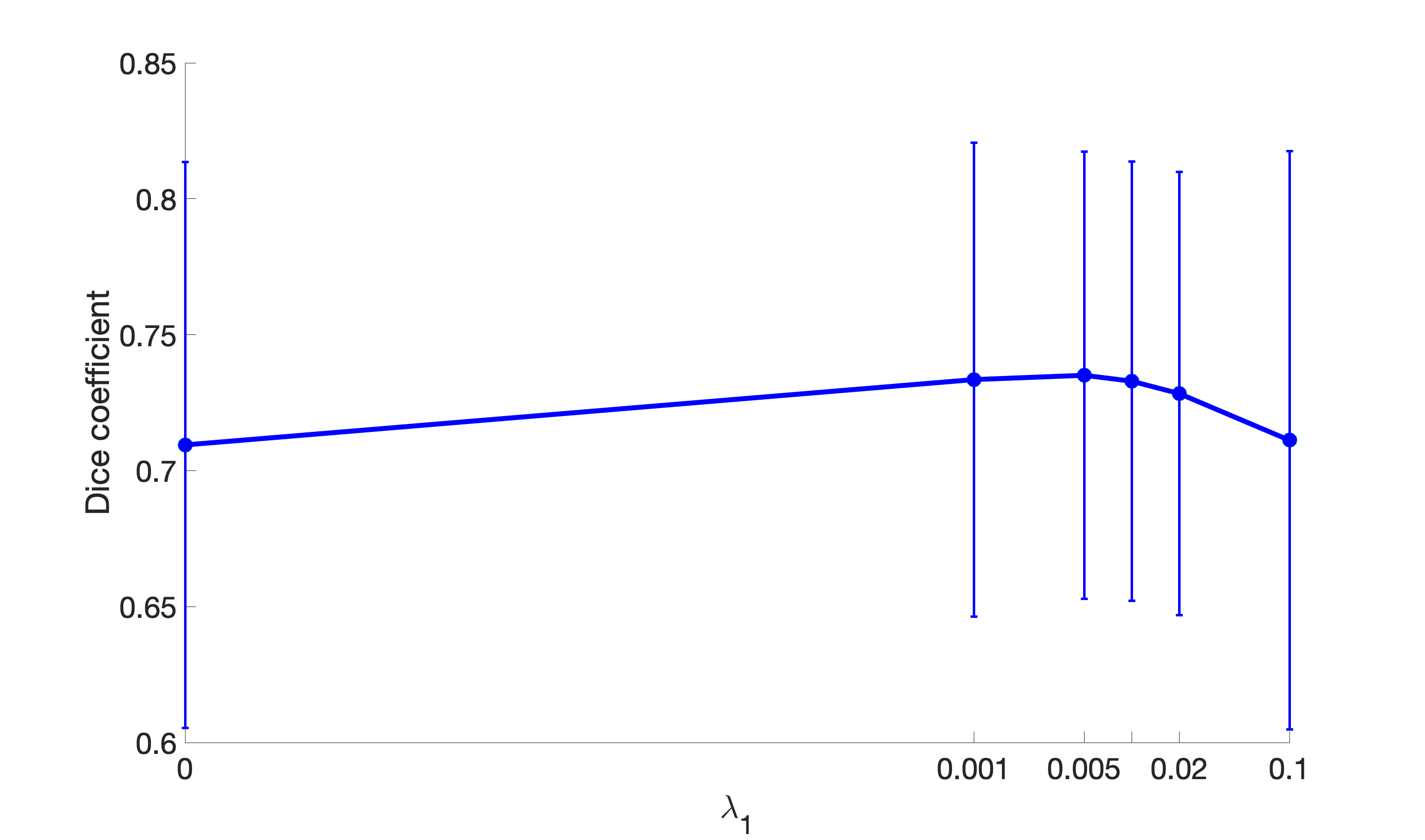}
\hfill
}
\vskip0.03in
\centerline{
\hfill
(b) Placentae: parameter sensitivity 
\hfill
}
\vskip0.08in
\centerline{
\hfill
\includegraphics[width=3.7in, height=2.1in]{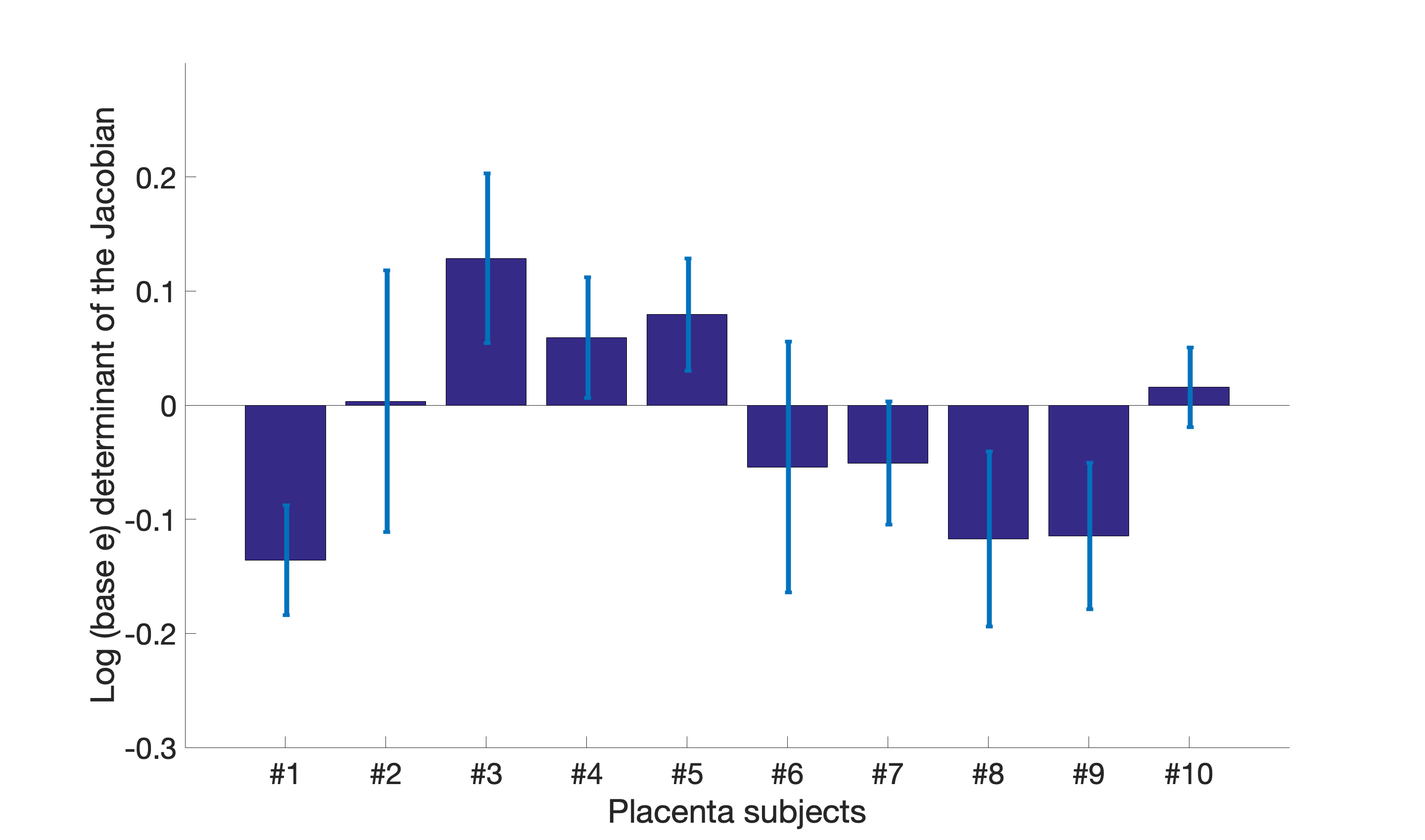}
\hfill
}
\vskip0.03in
\centerline{
\hfill
(c) Placentae: volumetric distortion
\hfill
}
\vskip0.08in
\caption{Volume overlap between propagated and manual placenta segmentations and volumetric distortion for the B-Spline transformations estimated by our temporal registration. Subfigure~(a): The 10 subjects in the study are reported in the increasing order of volume overlap of the original segmentation labels (no motion correction). Statistics are reported for our method (red), pairwise registration to the template frame (green), and no motion correction (blue). The volume overlap in temporal registration is $3.1\% ({\pm}7.9\%)$ higher than that in pairwise registration on average. Subfigure~(b): The volume overlap in the region of the placenta with B-Spline transformation model as a function of $\lambda_1$. Subfigure~(c): The mean and the standard deviation of the log determinants of the Jacobian matrix in each case (in total, 10 cases) are reported. All logs are base e. Thus, $\pm{0.7}$ approximately corresponds to expansion (or shrinking) by a factor of 2.}
\label{fig:dice_placenta}
\end{figure}

Fig.~\ref{fig:dice_placenta} reports volume overlap statistics for the placentae in this study. Our temporal alignment offers consistent improvement for placentae for all cases over pairwise registration to the reference frame. Fig.~\ref{fig:dice_placenta} also reports the sensitivity of the alignment accuracy with respect to the weight of temporal smoothness regularization in Eq.~(\ref{messagePassing2}). For B-Spline registration on placenta, the algorithm yields the best results when $\lambda_{1}={0.005}$. Finally, Fig.~\ref{fig:dice_placenta} shows the mean and the standard deviation of the log determinants in each case (in total, 10 cases). The mean values of the log determinants in each subject are all between $-0.15$ and $0.15$. A log determinant value above $0$ indicates volume expansion of the transformation, and a log determinant value below $0$ indicates volume compression. All the determinants of Jacabian of our estimated B-Spline transformations for placentae are positive.  

\section{Discussion}
\label{section:discussion}

Our temporal registration approach improves alignment of fetal brains and placentae in the in-utero MRI times series. The temporal model offers substantial improvement for fetal brain alignment in the presence of large motion, which presents significant challenges for the pairwise registration method. The incorporation of temporal smoothness improves the alignment of the placenta for every subject. In the study, the log determinants of Jacobian of the estimated B-Spline transformations are close to $0$, which indicates no over-expansion or over-compression in the resulting placenta transformation estimates. 

One important downstream use of this temporal registration method is modeling of ROI-specific intensity changes in the in-utero MRI time series~\cite{luo2017vivo, turk2017spatiotemporal}. Currently the signal dynamics in fetal brains, which tend to have a wide range of motion, include substantial fluctuations, due to spatial signal non-uniformities that affect the estimated signal if the fetus moves substantially in the volumetric image. The cross-correlation measure of image similarity used in our temporal registration captures local image differences, but tracking the intensity changes in corresponding regions across images is still affected by the signal non-uniformities. Future work will obtain robust estimates of the in-utero MRI signal time-courses by augmenting the temporal method with a model of ROI-specific intensity changes.

In addition, the registration of each frame to the template frame can be based on all the images in the time series by including a backward message pass. One of the challenges is to find a template that is close to both the first observation (image) and the last observation (image) in the HMM. Otherwise, estimating the first or the last hidden state (transformation) might fail due to the substantial difference between the moving image and the template image, since there is no temporal information passed to the very first step of the sequential estimation .

Our temporal registration method naturally accepts different types of transformation models besides the rigid model and the B-Spline model. For example, it can be extended to diffeomorphic transformation. The diffeomorphic demons algorithm~\cite{vercauteren2009diffeomorphic} adapts convolutional kernels on the incremental vector field and the updated displacement field at each iteration acts as a regularization in the cost function, which can naturally be implemented with our temporal registration method to incorporate temporal smoothness.

\section{Conclusions}
\label{section:conclusions}

We present a HMM-based registration method to align images in MRI time series. Forward message passing incorporates the temporal model of motion into the estimation procedure. The filtered estimates are therefore based on not only the present image frame and the template, but also on the previous frames in the image series. Our method can be easily extended on other forms of transformation models and image similarity metrics. The experiments on in-utero BOLD MRI time series demonstrate the promise of our approach in this novel and challenging application.

\end{document}